\documentclass[draftclsnofoot, 11pt, onecolumn]{IEEEtran}
\usepackage[left=1.25in,right=1.25in,top=1in,bottom=1in]{geometry}

\usepackage{tabularx,amsmath,amssymb,graphicx,paralist,subfigure,pdfpages,cite,algorithm,algpseudocode,paralist,multirow,comment}

\usepackage[autostyle]{csquotes}

\newtheorem{rmk}{Remark}

\def\ln{{\rm ln}}

\def\mc{\mathcal}
\def\mb{\mathbf}
\def\mbb{\mathbb}
\def\ra{\rightarrow}

\IEEEoverridecommandlockouts
\def\mbb{\mathbb}
\def\mb{\mathbf}
\def\mc{\mathcal}
\def\wh{\widehat}
\def\wt{\widetilde}
\def\ol{\overline}
\def\ul{\underline}

\def\bds{\boldsymbol}
\def\bth{\boldsymbol\theta}
\newcommand{\mn}[1]{{\left\vert\kern-0.25ex\left\vert\kern-0.25ex\left\vert\kern0.3ex #1 
		\kern0.3ex\right\vert\kern-0.25ex\right\vert\kern-0.25ex\right\vert}}

\begin{document}
	\title{\huge Gradient tracking and variance reduction for decentralized optimization and machine learning}
	\author{Ran Xin,  Soummya Kar, and Usman A. Khan
	}
	\maketitle

\vspace{-1.5cm}
\begin{abstract}
\vspace{-0.4cm}
Decentralized methods to solve finite-sum minimization problems are important in many signal processing and machine learning tasks where the data is distributed over a network of nodes and raw data sharing is not permitted due to privacy and/or resource constraints. In this article, we review decentralized stochastic first-order methods and provide a unified algorithmic framework that combines variance-reduction with  gradient tracking to achieve both robust performance and fast convergence. We provide explicit theoretical guarantees of the corresponding methods when the objective functions are smooth and strongly-convex, and show their applicability to non-convex problems via numerical experiments. Throughout the article, we provide intuitive illustrations of the main technical ideas by casting appropriate tradeoffs and comparisons among the methods of interest and by highlighting applications to decentralized training of machine learning models.  
\end{abstract}

	
\vspace{-0.5cm}
\section{Introduction}
In multi-agent networks and large-scale machine learning, when data is available at different devices with limited communication, it is often desirable to seek scalable learning methods that do not require bringing, storing, and processing data at one single location. In this article, we describe decentralized, stochastic first-order methods, which are particularly favorable to such ad-hoc and resource-constrained settings. Specifically, we describe a unified algorithmic framework for combining different \textbf{\textit{variance reduction methods}} with \textbf{\textit{gradient tracking}} in order to significantly improve upon the performance of the standard decentralized stochastic gradient descent (DSGD). However, this improvement comes at a price of losing the simplicity of DSGD and we study the added communication, computation, and storage requirements with the help of precise technical statements. For the ease of accessibility, we restrict the theoretical arguments to smooth and strongly-convex objectives, while the applicability to non-convex problems is shown with the help of numerical experiments.  We emphasize that smooth and strongly-convex objectives are relevant in many machine learning applications, e.g., problems where a strongly-convex regularization is added to otherwise convex costs, or problems where the objective functions are non-convex but strongly-convex in the neighborhood of the local minimizers~\cite{OPT_ML}. To provide context, we start by briefly reviewing the problems of interest and their associated centralized solutions.
	
\subsection{Empirical Risk Minimization}\label{sERM}
In parametric learning and inference problems, the goal of a typical machine learning system is to find a model~$g$, parameterized by a real vector~$\bds{\theta}\in\mathbb{R}^p$, that maps an input data point~$\mb{x}\in\mathbb{R}^{d_{\mb{x}}}$ to its corresponding output~$\mb{y}\in\mathbb{R}^{d_{\mb{y}}}$. The setup requires defining a loss function~$l(g(\bth;\mb{x,y}))$, which represents the loss incurred by the model~$g$ with parameter~$\bds{\theta}$ on the data~$(\mb{x},\mb{y})$. In the formulation of statistical machine learning, we assume that each data sample~$(\mb{x},\mb{y})$ belongs to a joint probability distribution~$\mc{P}(\mb{x},\mb{y})$. Ideally, we would like to find the optimal model parameter~$\widetilde{\bds{\theta}}^*$ by minimizing the following \textit{risk (expected loss) function}~$\widetilde{F}(\bds{\theta})$:
\begin{equation*}
\mbox{P0:} \qquad\widetilde{\bds{\theta}}^* = \operatorname*{argmin}_{\bds{\theta}\in\mathbb{R}^p}
\widetilde{F}(\bds{\theta}),\qquad \widetilde{F}(\bds{\theta}) \triangleq 
\mathbb{E}_{(\mb{x},\mb{y})\sim\mc{P}(\mb{x},\mb{y})}
l(g(\bth;\mb{x,y})).
\end{equation*} 
However, the true distribution~$\mc{P}(\mb{x},\mb{y})$ is often hidden or intractable in practice. In supervised machine learning, one usually has access to a large set of training samples~$\{\mb{x}_i,\mb{y}_i\}_{i=1}^N$, which can be considered as independent and identically distributed (i.i.d.) realizations from the distribution~$\mc{P}(\mb{x},\mb{y})$. The average of the losses incurred by the model~$\bds{\theta}$ on a finite set of training data samples~$\{\mb{x}_i,\mb{y}_i\}_{i=1}^N$, known as the \textit{empirical risk}, thus serves as an appropriate surrogate for the risk function~$\widetilde{F}(\bds{\theta})$. Formally, the \textit{empirical risk minimization} problem is stated as 
\begin{equation}
\mbox{P1:}\qquad\bds{\theta}^* = \operatorname*{argmin}_{\bds{\theta}\in\mathbb{R}^p}F(\bds{\theta}),\qquad F(\bds{\theta})\triangleq\frac{1}{N}\sum_{i=1}^{N}l(g(\bth;\mb{x}_i,\mb{y}_i)) \triangleq \frac{1}{N}\sum_{i=1}^{N} f_{i}(\bds{\theta}),
\end{equation}
where~$\bds\theta^*$ is the minimizer of the empirical risk~$F$. This finite-sum formulation captures a wide range of supervised learning problems. Examples include: hand-written character recognition with regularized logistic regression where the objective functions are smooth and strongly-convex~\cite{PRML}; text classification with support vector machines where the objectives are convex but not necessarily smooth~\cite{OPT_ML}; and perception tasks with deep neural networks where the cost functions are non-convex in general~\cite{OPT_ML,PRML}.

Our focus in this article is on smooth and strongly-convex objective functions defined as follows. An \textit{$L$-smooth} and \textit{$\mu$-strongly-convex} function~$f: \mathbb{R}^p\ra\mathbb{R}$ is such that~$\forall\bds{\theta}_1, \bds{\theta}_2\in\mbb{R}^p$ and for some positive constants~$L,\mu>0$, we have $$\frac{\mu}{2}\|\bds{\theta}_1-\bds{\theta}_2\|_2^2\leq f(\bds{\theta}_2)-f(\bds{\theta}_1)-\nabla f(\bds{\theta}_1)^\top(\bds{\theta}_2-\bds{\theta}_1)\leq\frac{L}{2}\|\bds{\theta}_1-\bds{\theta}_2\|_2^2.$$ We define~$\mc{S}_{\mu,L}$ as the class of functions that are~$L$-smooth and~$\mu$-strongly-convex~\cite{nesterov_book}. We note that if~$F\in\mc{S}_{\mu,L}$, then it has a unique global minimum denoted as~$\bth^*$. For any~$F\in\mc{S}_{\mu,L}$, we have that~$L\geq\mu$, and we define~$\kappa\triangleq\frac{L}{\mu}$ as the condition number of~$F$~\cite{nesterov_book}; clearly,~$\kappa\geq1$. For the ease of accessibility, we restrict the theoretical arguments to the function class~$\mc{S}_{\mu,L}$, while the applicability to non-convex  problems  is  shown  with  the  help  of numerical  experiments.

\subsection{Stochastic Gradient Descent}\label{sec_sgd}
Stochastic Gradient Descent (SGD) is a simple yet powerful method that has been extensively used to solve the empirical risk minimization problem P1. SGD, in its simplest form, starts with an arbitrary~$\bds{\theta}_0\in\mathbb{R}^p$ and performs the following iterations to learn~$\bth^*$ as~$k\ra\infty$:
\begin{align}\bds{\theta}_{k+1} = \bds{\theta}_k - \alpha_k \cdot  \nabla f_{s_k}(\bds{\theta}_k), \qquad k\geq0, \label{csgd}
\end{align}
where~$s_k$ is chosen uniformly at random from~$\{1,\cdots,N\}$ and~$\{\alpha_k\}_{k\geq 0}$ is a sequence of positive step-sizes. Comparing to batch gradient method where the descent direction~$\nabla F(\bth_k)$ at each iteration~$k$ is computed from the entire batch of data, SGD iteratively descends in the direction of the gradient of a randomly sampled component function. SGD is thus computationally-efficient as it evaluates one component gradient (extendable to more than one randomly selected functions) at each iteration and is a popular alternative in problems with a large number of high-dimensional training data samples and model parameters.

We note that the stochastic gradient~$\nabla f_{s_k}(\bth_k)$ is an unbiased estimate of batch gradient~$\nabla F(\bth_k)$, i.e.,~$\mathbb{E}_{s_k}[\nabla f_{s_k}(\bds{\theta}_k)|\bth_k] = \nabla F(\bth_k)$. Under the assumptions that each~$f_i\in\mc{S}_{\mu,L}$ and each stochastic gradient~$\nabla f_{s_k}(\bds\theta_k)$ has bounded variance\footnote{\label{f1}In this article, we restrict to the bounded variance assumption for simplicity. This assumption however can be relaxed, see~\cite{OPT_ML,NIPS2014_5355,gower2019sgd}, for example.}, i.e.,~$\mathbb{E}_{s_k}\left[\left\|\nabla f_{s_k}(\bds\theta_k)-\nabla F(\bds\theta_k)\right\|_2^2 |\bds\theta_k\right] \leq{\sigma}^2,\forall k,$ we note that with a constant step-size~$\alpha\in\left(0,\frac{1}{L}\right]$,~$\mathbb{E}\left[\|\bds{\theta}_k-\bds{\theta}^*\|_2^2\right]$ decays linearly (on the log-scale), at the rate of~$\left(1-\mu\alpha\right)^k$, to a neighborhood of~$\bth^*$. Formally, we have~\cite{OPT_ML},
\begin{align}\label{sgd_conv}
\mathbb{E}\left[\|\bds{\theta}_k-\bds{\theta}^*\|_2^2\right]
\leq (1-\mu\alpha)^k + \frac{\alpha{\sigma}^2}{\mu},\qquad \forall k\geq0.
\end{align}
This steady-state error $\frac{\alpha\sigma^2}{\mu}$ or the \textit{inexact convergence} is due to the fact that~$\nabla f_{s_k}(\bds{\theta}^*) \neq 0$, in general, and the step-size is constant. A diminishing step-size overcomes this~issue~and~leads~to~an \textit{exact convergence} to the minimizer~$\bds\theta^*$ albeit at slower rate. For example, with~$\alpha_k = \frac{1}{\mu (k+1)}$,~we~have
\begin{align}\label{sgd_conv_2}
\mathbb{E}\left[\|\bds{\theta}_k-\bds{\theta}^*\|_2^2\right]
\leq \frac{\max\left\{\frac{2{\sigma}^2}{\mu^2},\|\bth_0-\bth^*\|_2^2\right\}}{k+1},
\end{align}
for all $k\geq0$,~\cite{OPT_ML}. In other words, to reach an~$\epsilon$-accurate solution of~$\bth^*$,~i.e.,~$\mathbb{E}\left[\|\bds{\theta}_k-\bds{\theta}^*\|^2\right]\leq\epsilon$, SGD (with decaying step-sizes) requires~$\mc{O}\left(\frac{1}{\epsilon}\right)$ component gradient evaluations. 

\subsection{Variance-Reduced Stochastic Gradient Descent}\label{VRSGD}
In practice, a successful implementation of SGD relies heavily on the tuning of the step-sizes, and typically a decaying step-size sequence~$\{\alpha_k\}_{k\geq0}$ has to be carefully chosen due to the potentially large variance in SGD, i.e., the sampled gradient~$\nabla f_{s_k}(\bds{\theta}_k)$ at~$\bds{\theta}_k$ can be very far from the batch gradient~$\nabla F(\bds{\theta}_k)$. In recent years, certain Variance-Reduction (VR) techniques have been developed towards addressing this issue~\cite{SAG,SAGA,SVRG,SARAH}. The key idea here is to design an iterative estimator of the batch gradient whose variance progressively decays to zero as~$\bth_k$ approaches~$\bth^*$. Benefiting from this reduction in variance, VR methods have a low per-iteration computation cost, a key feature of SGD, and, at the same time, converge linearly to the minimizer~$\bth^*$ as the batch gradient descent (with a \textit{constant} step-size for the objective function class~$\mc{S}_{\mu,L}$). Different constructions of the aforementioned gradient estimator lead to different VR methods~\cite{SAG,SAGA,SVRG,SARAH}. We focus on two popular VR methods in this article described as follows.

\textbf{SAGA~\cite{SAGA}: }The SAGA method starts with an arbitrary~$\bth_0\in\mathbb{R}^p$ and maintains a table that stores all component gradients~$\{\nabla f_i(\wh{\bth}_i)\}_{i=1}^N$, where~$\wh{\bth}_{i}$ denotes the most~recent iterate at which~$\nabla f_i$ was evaluated, initialized with~$\{\nabla f_i(\bth_0)\}_{i=1}^N$. At every iteration~$k\geq0$, SAGA chooses an index~$s_k$ uniformly at random from~$\{1,\ldots,N\}$ and performs the following two updates:
\begin{align}
    \mb{g}_k = \nabla f_{s_k}(\bth_k) - \nabla f_{s_k}(\wh{\bth}_{s_k}) + \frac{1}{N}\sum_{i=1}^N\nabla f_i(\wh{\bth}_{i}),\qquad \qquad \bth_{k+1}=\bth_k-\alpha\cdot \mb{g}_k.
\end{align}
Subsequently, the entry~$\nabla   f_{s_k}(\wh{\bth}_{s_k})$ in the gradient table is replaced by~$\nabla f_{s_k}\big(\bth_{k}\big)$, while the other entries remain unchanged. Under the assumption that each~$f_i\in\mc{S}_{\mu,L}$, it can be shown that with~$\alpha = \frac{1}{3L}$, we have~\cite{SAGA},
\begin{align}
\mathbb{E}\left[\left\|\bth_k-\bth^*\right\|_2^2\right]\leq C\left(1-\min\left\{\frac{1}{4N},\frac{1}{3\kappa}\right\}\right)^k,\qquad \forall k\geq0,
\end{align}
for some~$C>0$. In other words, SAGA achieves~$\epsilon$-accuracy of~$\bth^*$ with~$\mc{O}\left(\max\{N,\kappa\}\log\frac{1}{\epsilon}\right)$ component gradient evaluations, where recall that~$\kappa=\frac{L}{\mu}$ is the condition number of the global objective function~$F$. 
Indeed, SAGA has a non-trivial storage cost of~$\mc{O}\left(Np\right)$ due to the gradient table, which can be reduced to~$\mc{O}(N)$ for certain problems of interest, for example, logistic regression and least squares, by exploiting the structure of the objective functions~\cite{SAG,SAGA}. 

\textbf{SVRG~\cite{SVRG}: }Instead of storing the gradient table, SVRG achieves variance reduction by computing the batch gradient periodically and can be interpreted as a \textit{double-loop} method described as follows. The outer loop of SVRG, indexed by~$k$, updates the estimate~$\bth_k$ of~$\bth^*$. At each outer iteration~$k$, SVRG computes the batch gradient~$\nabla F(\bds{\theta}_k)$ and executes a finite number~$T$ of SGD-type inner loop iterations, indexed by~$t$: with~$\ul\bth_{0} = \bth_k$ and for~$t = 0,\cdots,T-1$, 
\vspace{-.1cm}
\begin{align}\label{csvrg}
\mb{v}_{t} = \nabla f_{s_{t}}(\ul\bth_{t}) - \nabla f_{s_{t}}(\ul\bth_{0}) + \nabla F(\ul\bth_{0}), \qquad\qquad
\ul\bth_{t+1} = \ul\bth_{t} - \alpha \cdot \mb{v}_{t}, 
\end{align}
where the index~$s_{t}$ is selected uniformly at random from~$\{1,\cdots,N\}$. After the inner loop completes, $\bds{\theta}_{k+1}$ can be updated in a few different ways; applicable choices include setting~$\bds{\theta}_{k+1}$ as~$\ul\bth_{T}$,~$\tfrac{1}{T}\sum_{t=0}^{T-1}\ul\bth_{t}$, or choosing it randomly from the inner loop updates~$\{\ul\bth_{t}\}_{t=0}^{T-1}$. For instance, assuming that each~$f_i\in\mc{S}_{\mu,L}$, it can be shown that with~$\bds{\theta}_{k+1}=\tfrac{1}{T}\sum_{t=0}^{T-1}\ul\bth_{t}$,~$\alpha = \frac{1}{10L}$, and~$T = 50\kappa$, we have~\cite{SVRG},~$$
\mathbb{E}[\|\bds{\theta}_{k}-\bth^*\|^2]\leq D \cdot 0.5^k,\qquad \forall k\geq 0,$$
for some~$D>0$. That is to say, SVRG achieves~$\epsilon$-accuracy with~$\mc{O}(\log\frac{1}{\epsilon})$ outer-loop iterations. We further note that each outer-loop update requires~$N+2T$ component gradient evaluations~\eqref{csvrg}. Therefore, SVRG achieves~$\epsilon$-accuracy of~$\bth^*$ with~$\mc{O}\left((N+\kappa)\log\frac{1}{\epsilon}\right)$ component gradient evaluations, which is comparable to the convergence rate of SAGA.

\begin{rmk}[SGD with decaying step-sizes vs. VR]\label{VRdisc}
SGD, converging at a sublinear rate~$\mc O(1/k)$ to the minimizer~\eqref{sgd_conv_2}, typically makes a fast progress in its early stage for certain large-scale, complex machine learning tasks and then slows down considerably. Its complexity~\eqref{sgd_conv_2} is not explicitly dependent on the sample size~$N$, which is a strong feature, but it comes at a price of a direct dependence on~$\sigma^2$ (the variance of the stochastic gradient). On the other hand, the VR methods achieve fast linear convergence with the help of refined gradient estimators, for example,~$\mb{g}_k$ or~$\mb{v}_{t}$, which approach the corresponding batch gradients as their variance diminishes. Their convergence, although dependent on the sample size~$N$, is independent of~$\sigma^2$. 

\end{rmk}

\begin{rmk}[SAGA vs. SVRG]\label{SAGA_SVRG}
The fundamental trade-off between SAGA and SVRG is convergence speed versus storage and is often described as a trade-off between time and space~\cite{SAGA}. Although SAGA and SVRG in theory achieve convergence rates of the same order, SVRG in practice requires~$2$-$3$ times more component gradient evaluations to reach the same accuracy as SAGA, however, without storing all the component gradients~\cite{SAGA}.
\end{rmk}

In the rest of this article, we show how to cast SGD and VR methods in the decentralized optimization framework. \textit{Section: Problem Formulation} describes the decentralized optimization problem over a network of nodes. In \textit{Section: Decentralized Stochastic Optimization}, we extend centralized SGD to the decentralized problem and show that an appropriate decentralization is achieved with the help of gradient tracking. Subsequently, in \textit{Section: Decentralized VR Methods}, we describe recent advances in decentralized methods that combine gradient tracking and variance reduction. \textit{Section: Numerical Illustrations} provides numerical experiments on strongly-convex and non-convex problems and further highlights different tradeoffs between the methods described in this article. \textit{Section: Extensions and Discussion} summarizes certain extensions and communication/computation aspects of the corresponding problems that are popular in the literature. Finally, \textit{Section: Conclusions} concludes the paper and briefly describe some open problems.

\section{Problem Formulation: Decentralized Empirical Risk Minimization}\label{secPF}
In this article, our focus is on the solutions for optimization problems that arise in peer-to-peer decentralized networks. Unlike traditional master-worker architectures, where a central node acts as a master that coordinates communication with all workers; there is no  central coordinator in peer-to-peer networks and each node is only able to communicate with its immediate neighbors, see Fig.~\ref{decentralized}. The canonical form of decentralized optimization problems can be described as follows. Consider~$n$ nodes, such as machines, devices, or robots, that communicate over a static undirected graph~$\mc{G}=(\mc{V},\mc{E})$, where~$\mc{V}=\{1,\cdots,n\}$ is the set of nodes, and~$\mc{E}\subseteq\mc{V}\times\mc{V}$ is the set of edges, i.e., a collection of ordered pairs~$(i,r),i,r\in\mc{V}$, such that nodes~$i$ and~$r$ can exchange information. Following the discussion in \textit{Section: Empirical Risk Minimization}, each node~$i$ holds a \textit{local risk function},~$\wt f_i:\mathbb{R}^p\ra\mathbb{R}$, not accessible by any other node in the network. 
The decentralized risk minimization problem can thus be defined as 
\begin{align}
\mbox{P2:}
\quad\widetilde\bth^* = \operatorname*{argmin}_{\bds{\theta}\in\mathbb{R}^p}\wt F(\bds{\theta}),\qquad \wt F(\bds\theta)\triangleq\frac{1}{n}\sum_{i=1}^n\wt f_i(\bds\theta).\nonumber
\end{align}

As in the centralized case with Problem P0, the underlying data distributions at the nodes may not be available or tractable, we thus employ a local empirical risk function at each node as a surrogate of the local risk. Specifically, we consider each node~$i$ as a computing resource that stores/collects a local batch of~$m_i$ training samples that are possibly private (not shared with other nodes) and the corresponding local empirical risk function is decomposed over the local data samples as~$f_i\triangleq\frac{1}{m_i}\sum_{j=1}^{m_i}f_{i,j}$. The goal of the networked nodes is to \textit{agree} on the \textit{optimal} solution of the
following \textit{decentralized empirical risk minimization} problem:
\begin{align}
\mbox{P3:}
\quad\bds{\theta}^* = \operatorname*{argmin}_{\bds{\theta}\in\mathbb{R}^p}F(\bds{\theta}),\qquad F(\bds\theta)=\frac{1}{n}\sum_{i=1}^nf_i(\bds\theta)\triangleq\frac{1}{n}\sum_{i=1}^n\:\left(\frac{1}{m_i}\sum_{j=1}^{m_i}f_{i,j}(\bds\theta)\right).\nonumber
\end{align}
The rest of this article is dedicated to the solutions of Problem~P3. 
\begin{figure*}[!h]
\centering
\subfigure{\includegraphics[width=1.5in]{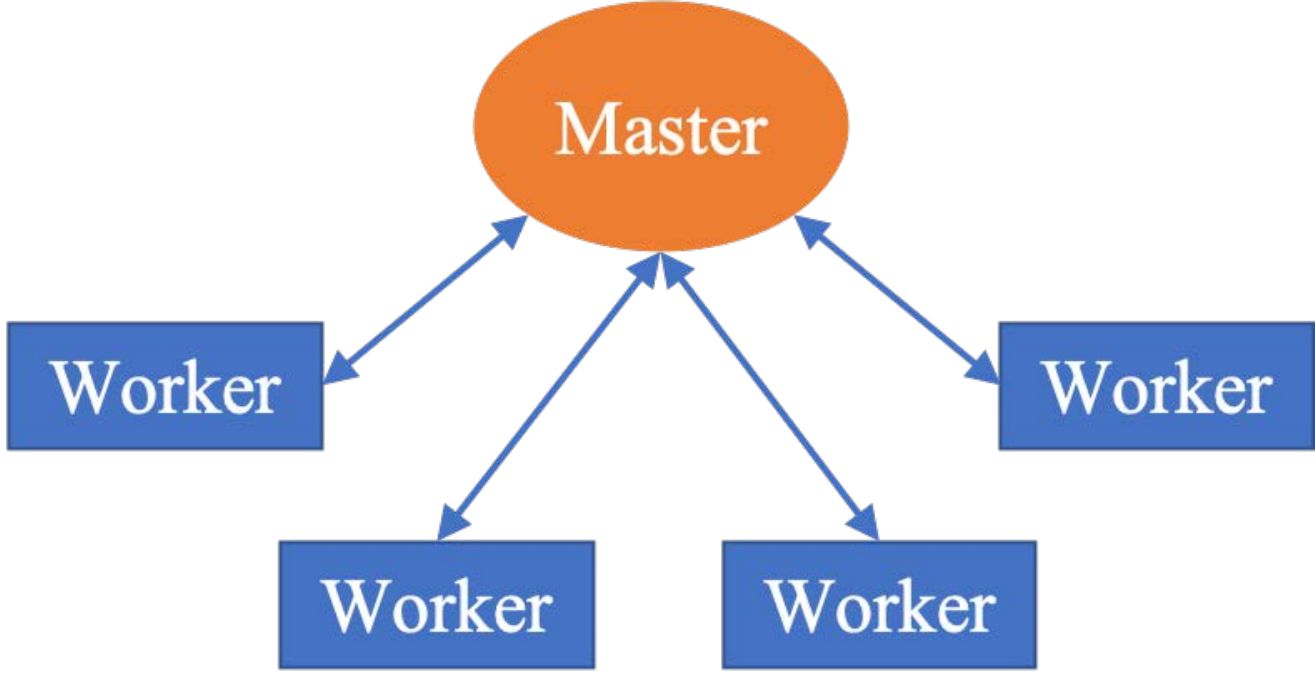}}\hspace{2cm}
\subfigure{\includegraphics[width=1.5in]{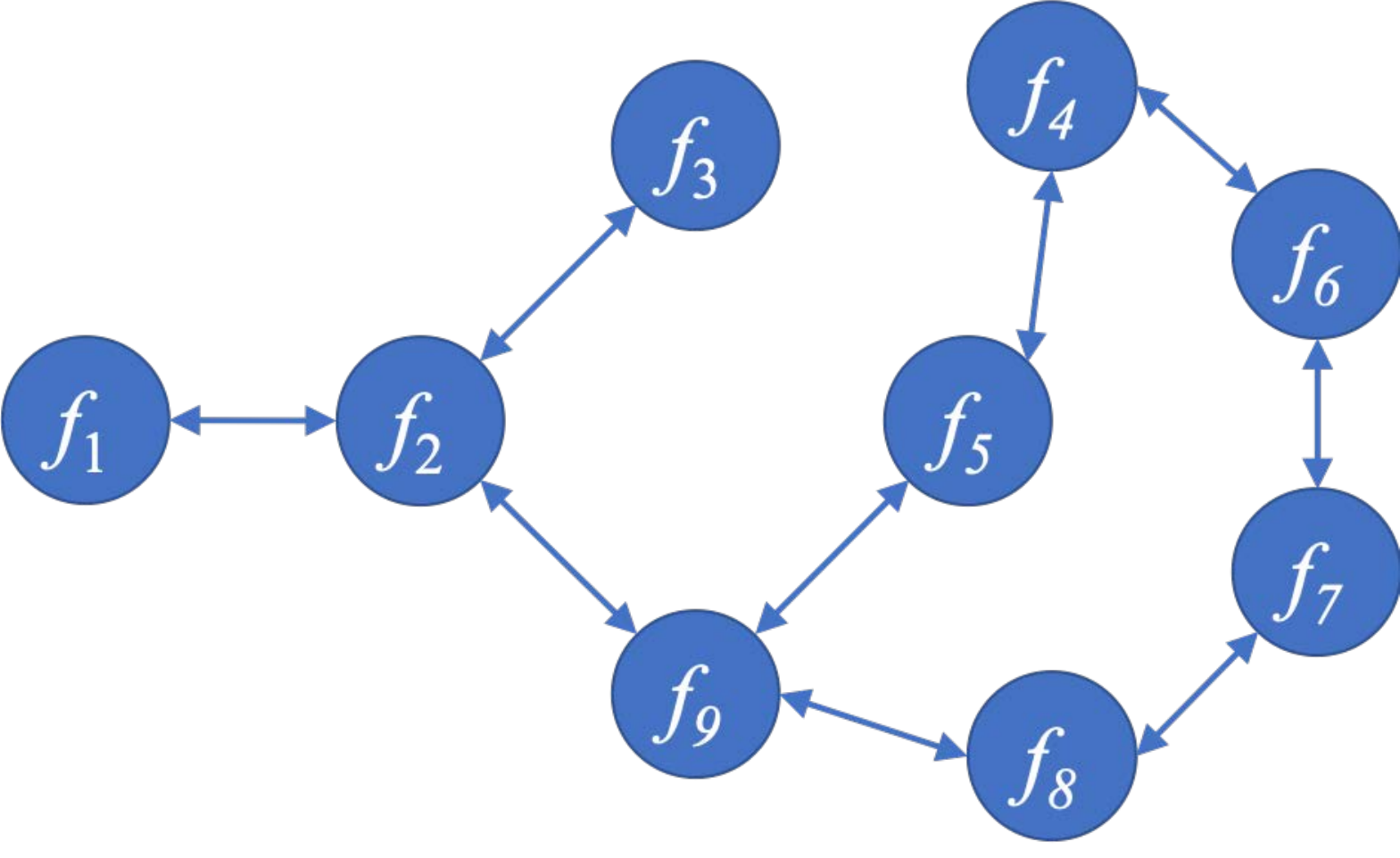}}
\caption{(Left) A master-worker network. (Right) Decentralized optimization in peer-to-peer networks.}
\label{master_worker}\label{decentralized}
\end{figure*}

\vspace{-0.4cm}
\section{Decentralized Stochastic Optimization}\label{S2}
We now consider decentralized solutions of Problem P3. At each node~$i$ given the current estimate~$\bth_k^i$ of~$\bth^*$ at iteration~$k$, related algorithms typically involve the following steps:
\begin{enumerate}
\item Sample one or more component gradients from~$\{\nabla f_{i,j}(\bth_k^i)\}_{j=1}^{m_i}$;
\item Fuse information with the available neighbors;
\item Compute~$\bds\theta_{k+1}^i$ according to a specific optimization protocol.
\end{enumerate}
Recall that each node in the network only communicates with a few nearby nodes and only has partial knowledge of the global objective, see Fig.~\ref{decentralized} (right). Due to this limitation, an information propagation mechanism is required that disseminates local information over the entire network. Decentralized optimization thus has two key components: (i) \textit{agreement or consensus}--all nodes must agree on the same state, i.e.,~$\bds\theta_k^i\ra\bds\theta_{cons},\forall i$; and, (ii) \textit{optimality}--the agreement should be on the minimizer of the global objective~$F$, i.e.,~$\bds\theta_{cons} = \bds\theta^*$. Average-consensus algorithms~\cite{consensus_Murray} are information fusion protocols that enable each node to appropriately combine the vectors received from its neighbors and to agree on the average of the initial states of the nodes. They thus naturally serve as basic building blocks in decentralized optimization, added to which are local gradient corrections that steer the agreement to the global minimizer. 

To describe average-consensus, we first associate the undirected and connected graph~$\mc{G}$ with a primitive, symmetric, and doubly-stochastic~$n\times n$ weight matrix~$W=\{w_{ir}\}$, such that~$w_{ir}\neq0$ for each~$(i,r)\in\mc{E}$. Clearly, we have~$W=W^\top$ and~$W\mb{1}_n=\mb{1}_n$, where~$\mb 1_n$ is the column vector of~$n$ ones. There are various ways of constructing such weights in a decentralized manner. Popular choices include the Laplacian and Metropolis weights, see~\cite{tutorial_nedich} for details. Average-consensus~\cite{consensus_Murray} is given as follows. Each node~$i$ starts with some vector~$\bth_0^i\in\mathbb{R}^p$ and updates its state according to~$\bth_{k+1}^i = \sum_{r\in\mc{N}_i}w_{ir} \bth_{k}^r$,~$\forall k\geq0$. It can be written in a vector form as
\begin{equation}\label{average_consensus_undirected_matrix}
\bds\theta_{k+1} = (W\otimes I_p) \bds\theta_{k},
\end{equation} 
where~$\bds\theta_k=[{\bth_k^1}^\top,\cdots,{\bth_k^n}^\top]^\top$. Since~$W$ is primitive and doubly-stochastic\footnote{In the rest of this article,~$W=\{w_{ir}\}$ denotes a collection of doubly-stochastic weights and~$\lambda\in[0,1)$ is the spectral radius of~$(W-\frac{1}{n}\mb{1}_n\mb{1}_n^\top)$.}, from the Perron-Frobenius theorem~\cite{matrix_analysis}, we have~$\lim\limits_{k\ra\infty} W^k = \frac{1}{n}\mb{1}_n\mb{1}_n^\top$ and~$
\lim\limits_{k\ra\infty}\bds\theta_{k} = (W\otimes I_p)^k\bds\theta_0 = (\mb{1}_n\otimes I_p) \ol{\bth}_0,$
where~$\ol{\bth}_0 \triangleq \frac{(\mb{1}_n^\top\otimes I_p)\bth_0}{n}$, at a linear rate of~$\lambda^k$, and~$\lambda\in[0,1)$ is the spectral radius of~$(W-\frac{1}{n}\mb{1}_n\mb{1}_n^\top)$. That is to say, the protocol in~\eqref{average_consensus_undirected_matrix} enables an agreement across the nodes on the average~$\ol{\bth}_0$ of their initial states, at a linear rate. With the agreement protocol in place, we next introduce decentralized gradient descent and its stochastic variant that build on top of average-consensus.
	
\subsection{Decentralized Stochastic Gradient Descent (DSGD)}
Recall that our focus is to solve Problem P3 in a decentralized manner, when the nodes exchange information over an arbitrary undirected graph.
A well-known solution to this problem is Decentralized Gradient Descent (DGD)~\cite{DGD_nedich,diffusion_Chen}, described as follows. Each node~$i$ starts with an arbitrary~$\bth_0^i\in\mathbb{R}^p$ and performs the following update:
\begin{equation}\label{DGD}
\bds\theta_{k+1}^i = \sum_{r\in\mc{N}_i}w_{ir}\bds\theta_{k}^r - \alpha_k\nabla f_i\left(\bds\theta_k^i\right),\qquad k\geq 0.
\end{equation}
Indeed, at each node~$i$, DGD adds a local gradient correction to average-consensus based on the local data batch, i.e., all~$f_{i,j}$'s, and is the prototype of many decentralized optimization protocols. To understand the iterations of DGD, we write them in a vector form. Let~$\bds\theta_k$ and~$\nabla\mb{f}(\bds\theta_k)$ collect all local estimates and gradients, respectively, i.e.,~$\bds\theta_k=[{\bds\theta_k^1}^\top,\cdots,{\bds\theta_k^n}^\top]^\top$ and~$\nabla\mb{f}(\bds\theta_{k})\triangleq[{\nabla f_1(\bds\theta_k^1)}^\top,\cdots,{\nabla f_n(\bds\theta_k^n)}^\top]^\top$, both in~$\mathbb{R}^{np}$. Then DGD can be compactly written as
\begin{equation}\label{DGD_matrix}
\bds\theta_{k+1} = (W\otimes I_p)\bds\theta_k - \alpha_k\nabla\mb{f}(\bds\theta_k).
\end{equation}
We further define the average~$\ol{\bds\theta}_k \triangleq \frac{1}{n}(\mb{1}_n^\top\otimes I_p)\bds\theta_k$ of the local estimates at time~$k$ and multiply both sides of~\eqref{DGD_matrix} by~$\frac{1}{n}(\mb{1}_n^\top\otimes I_p)$ to obtain:
\begin{equation}\label{DGD_average}
\ol{\bds\theta}_{k+1} = \ol{\bds\theta}_k - \alpha_k\frac{(\mb{1}_n^\top\otimes I_p)\nabla\mb{f}\left(\bds\theta_k\right)}{n}.
\end{equation} 
Based on~\eqref{DGD_matrix} and~\eqref{DGD_average}, we note that the consensus matrix~$W$ makes the estimates~$\{\bds\theta_k^i\}_{i=1}^n$ at the nodes approach their average~$\ol{\bds\theta}_k$, while the average gradient~$\frac{(\mb{1}_n^\top\otimes I_p)\nabla\mb{f}\left(\bds\theta_k\right)}{n}$ steers~$\ol{\bds\theta}_k$ towards the minimizer~$\bth^*$ of~$F$. The overall protocol thus ensures agreement and optimality, the two key components of decentralized optimization as we described before. 

DGD is a simple yet effective method for various decentralized learning tasks. To make DGD efficient for large-scale decentralized empirical risk minimization, where each~$m_i$ is very large, Refs.~\cite{DSGD_nedich,diffusion_Chen} derive a stochastic variant, known as \textit{Decentralized Stochastic Gradient Descent (DSGD)}, by substituting each local batch gradient with a randomly sampled component gradient. DSGD is formally described in Algorithm 1. Assuming that each~$f_{i,j}\in\mc{S}_{\mu,L}$ and each local stochastic gradient has bounded variance\footnote{The bounded variance assumption can also be relaxed as noted in Footnote~\ref{f1} for the centralized case, see~\cite{DSGD_Yuan}.}, i.e.,~$\mathbb{E}_{s_k^i}\left[\left\|\nabla f_{i,s_k^i}(\bds\theta_k^i)-\nabla f_i(\bds\theta_k^i)\right\|_2^2 |\bds\theta_k^i\right] \leq\sigma^2,\forall i,k$, we have~\cite{DSGD_Yuan}: under a constant step-size,~$\alpha_k=\alpha\in\left(0,\mc{O}\left(\tfrac{(1-\lambda)}{L\kappa}\right)\right],\forall k$, $\mathbb{E}[\|\bds\theta_k^i-\bds\theta^*\|_2^2]$ decays at a linear rate of~$\left(1-\mc{O}(\mu\alpha)\right)^k$ to a neighborhood of~$\bth^*$ such that 
\begin{equation}\label{DSGD_convergence}
\limsup\limits_{k\ra\infty}\frac{1}{n}\sum_{i=1}^{n}\mathbb{E}\left[\left\|\bds\theta_k^i-\bds\theta^*\right\|_2^2\right]
=  \mc{O}\left(\frac{\alpha\sigma^2}{n\mu}
+ \frac{\alpha^2\kappa^2\sigma^2}{1-\lambda}
+ \frac{\alpha^2\kappa^2b}{\left(1-\lambda\right)^2}\right),
\end{equation}
where~$b \triangleq \frac{1}{n}\sum_{i=1}^{n}\left\|\nabla f_i\left(\bds\theta^*\right)\right\|^2$ and~$\kappa = L/\mu$. With a diminishing step-size~$\alpha_k = \mc{O}(\frac{1}{k})$, DSGD achieves an exact convergence~\cite{DSGD_Anit,DSGD_pu}, such that
\begin{align}\label{DSGD_diminishing}
\frac{1}{n}\sum_{i=1}^{n}\mathbb{E}\left[\left\|\bds\theta_k^i-\bds\theta^*\right\|_2^2\right] = \mc{O}\left(\frac{1}{k}\right),\qquad\forall k\geq 0.
\end{align}

\begin{algorithm}[!h]
		\caption{DSGD: At each node~$i$}
		\begin{algorithmic}[1]
			\Require $\bds\theta_0^i$,~$\{\alpha_k\}_{k\geq0}$,~$\{w_{ir}\}_{r\in\mc N_i}$. 
			\For{$k= 0,1,2,\cdots$}
			\State\textbf{Choose}~$s_k^i$ uniformly at random in~$\{1,\cdots,m_i\}$ 
			\State \textbf{Compute} the local stochastic gradient~$\nabla f_{i,s_k^i}(\bth_k^i)$.
			\State \textbf{Update}:~$\bds\theta_{k+1}^i = \sum_{r\in\mc{N}_i}w_{ir}\bds\theta_k^r
			- \alpha_k\nabla f_{i,s_k^i}(\bth_k^i)$
			\EndFor
		\end{algorithmic}
	\end{algorithm} 

\begin{rmk}[SGD vs. DSGD]\label{R2}
Comparing~\eqref{sgd_conv} to~\eqref{DSGD_convergence}, when a constant step-size~$\alpha$ is used, the steady-state error in both SGD and DSGD decays linearly to a certain neighborhood (controlled by~$\alpha$) of~$\bth^*$. Unlike SGD, however, the steady-state error of DSGD has an additional bias, independent of the variance~$\sigma^2$ of the stochastic gradient, that comes from~$b=\frac{1}{n}\sum_{i=1}^{n}\left\|\nabla f_i\left(\bds\theta^*\right)\right\|^2$. The constant~$b$ is not zero in general and characterizes the difference between the minimizer of each local objective~$f_i$ and that of the global objective~$F$. The resulting  bias~$\mc{O}\big(\frac{\alpha^2\kappa^2b}{\left(1-\lambda\right)^2}\big)$ can be significantly large when the data distributions across nodes are substantially heterogeneous or when the graph is not well-connected, a scenario that commonly arises in certain wireless networks and IoT applications, see \textit{Section: Numerical Illustrations}. In the following, we describe a gradient tracking technique that eliminates the bias in DSGD due to the term~$b$ and thus can be considered as a more appropriate decentralization of the centralized SGD.
\end{rmk}

\subsection{Decentralized First-Order Methods with Gradient Tracking}
\label{GT-DSGD}
To present the intuition behind the gradient tracking technique, we first recall the iterations of the (non-stochastic) Decentralized Gradient Descent (DGD) with a constant step-size in~\eqref{DGD}. Let us first assume, for the sake of argument, that all nodes agree on the minimizer of~$F$ at some iteration~$k$, i.e.,~$\bds\theta_k^i = \bds\theta^*,\forall i$. Then at the next iteration~$k+1$, we have
\begin{equation}\label{DGD_limit}
	\bds\theta_{k+1}^i = \sum_{r\in\mc N_i}w_{ir}\bds\theta^*-\alpha\nabla f_i(\bds\theta^*)=\bds\theta^*-\alpha\nabla f_i(\bds\theta^*),
\end{equation}
where~$\bds\theta^*-\nabla f_i(\bds\theta^*) \neq\bds\theta^*$, in general. In other words, the minimizer~$\bds\theta^*$ is not necessarily a fixed point of~\eqref{DGD}. Of course, using the gradient~$\nabla F\left(\bds\theta_k^i\right)$ of the \textit{global} objective, instead of~$\nabla f_i\left(\bds\theta_k^i\right)$, overcomes this issue but the global gradient is not available at any node. The natural yet innovative idea of gradient tracking is to design a local iterative gradient tracker~$\mb{d}_k^i$ that asymptotically approaches the global gradient~$\nabla F\left(\bds\theta_k^i\right)$ as~$\bds\theta_k^i$ approaches~$\bth^*$
\cite{NEXT_scutari,GT_CDC,GT_Qu,DIGing,AB}. Gradient tracking is implemented with the help of dynamic average consensus (DAC)~\cite{DAC}, briefly described next.
	
In contrast to classical average-consensus~\cite{consensus_Murray} that learns the average of fixed initial states, DAC~\cite{DAC} tracks the average of time-varying signals. Formally, each node~$i$ measures a time-varying signal~$\mb{r}_k^i$ and all nodes cooperate to track the average~$\ol{\mb{r}}_k\triangleq\frac{1}{n}\sum_{i=1}^{n}\mb{r}_k^i$ of these signals. The DAC protocol is given as follows. Each node~$i$ iteratively updates its estimate~$\mb{d}_k^i$ of~$\ol{\mb{r}}_k$ as
\begin{equation}\label{DAC}
\mb{d}_{k+1}^i = \sum_{r\in\mc{N}_i}w_{ir}\mb{d}_{k}^r + \mb{r}_{k+1}^i - \mb{r}_{k}^i, \qquad k\geq0,
\end{equation}
where~$\mb{d}_0^i = \mb{r}_{0}^i, \forall i$. For a doubly-stochastic weight matrix~$W=\{w_{ir}\}$, it is shown in~\cite{DAC} that if~$\left\|\mb{r}_{k+1}^i - \mb{r}_{k}^i\right\|_2\ra0$, we have that~$\left\|\mb{d}_{k}^i-\ol{\mb{r}}_k\right\|_2\ra0$. Clearly, in the aforementioned design of gradient tracking, the time-varying signal that we intend to track is the average of the local gradients~$\frac{1}{n}\sum_{i=1}^{n}\nabla f_i\left(\bds\theta_k^i\right)$. We thus combine DGD~\eqref{DGD} and DAC~\eqref{DAC} to obtain \textit{GT-DGD (DGD with Gradient Tracking)}~\cite{NEXT_scutari,GT_CDC,GT_Qu,DIGing,AB}, as follows:
	\begin{subequations}\label{DGT}
		\begin{align}
		\bds\theta_{k+1}^i &= \sum_{r\in\mc{N}_i}w_{ir}\bds\theta_{k}^r - \alpha\cdot\mb{d}_{k}^i, \label{DGT1}\\
		\mb{d}_{k+1}^i &= \sum_{r\in\mc{N}_i}w_{ir}\mb{d}_{k}^r + \nabla f_i\left(\bds\theta_{k+1}^i\right) - \nabla f_i\left(\bds\theta_k^i\right),\label{DGT2}
		\end{align}
	\end{subequations}
where~$\mb{d}_0^i = \nabla f_i\left(\bds\theta_0^i\right),\forall i$. Intuitively, as~$\bds\theta_k^i\ra\ol{\bds\theta}_k$ and~$\mb{d}_k^i\ra\frac{1}{n}\sum_{i=1}^n\nabla f_i\big(\bth_k^i\big)\ra\nabla F\big(\ol{\bth}_k\big)$,~\eqref{DGT1} asymptotically becomes the centralized batch gradient descent. It has been shown in~\cite{GT_Qu,DIGing,AB,DGT_NIPS} that GT-DGD converges linearly to the minimizer~$\bth^*$ of~$F$ under a~\textit{constant step-size} when each~$f_{i,j}\in\mc{S}_{\mu,L}$, unlike DGD that converges sublinearly to~$\bds\theta^*$ with decaying step-sizes.

The stochastic variant of GT-DGD is derived in~\cite{DSGT}, termed as~\textit{GT-DSGD (DSGD with Gradient Tracking)}, and is formally described in Algorithm~2. Under the same assumptions of smoothness, strong-convexity, and bounded variance as in DSGD, the convergence of GT-DSGD is summarized in the following~\cite{DSGT}: with a constant step-size,~$\alpha_k=\alpha\in\left(0,\mc{O}\left(\frac{(1-\lambda)^2}{L\kappa}\right)\right],\forall k$,~$\mathbb{E}[\|\bds\theta_k^i-\bds\theta^*\|_2^2]$ decays linearly at the rate of~$\left(1-\mc{O}(\mu\alpha)\right)^k$ to a neighborhood of~$\bth^*$ such that
\begin{equation}\label{DSGT_convergence}
	\limsup\limits_{k\ra\infty}\frac{1}{n}\sum_{i=1}^{n}\mathbb{E}\left[\left\|\bds\theta_k^i-\bds\theta^*\right\|_2^2\right]
	= \mc{O}\left(\frac{\alpha\sigma^2}{n\mu}
	+ \frac{\alpha^2\sigma^2\kappa^2}{\left(1-\lambda\right)^3}\right).
\end{equation}
Note that GT-DSGD, in contrast to GT-DGD, loses the exact linear convergence to the minimizer because the gradients are now stochastic. Exact convergence can be recovered albeit at a slower sublinear rate, i.e., with a diminishing step-size~$\alpha_k = \mc{O}(\frac{1}{k})$, we have~\cite{DSGT}
\begin{align}\label{DSGT_diminishing}
\frac{1}{n}\sum_{i=1}^{n}\mathbb{E}\left[\left\|\bds\theta_k^i-\bds\theta^*\right\|_2^2\right] = \mc{O}\left(\frac{1}{k}\right),\qquad\forall k\geq0.
\end{align} 
\begin{algorithm}[!h]
	\caption{GT-DSGD: At each node~$i$}
	\begin{algorithmic}[1]
		\Require$\bds\theta_0^i$,~$\{\alpha_k\}_{k\geq0}$,~$\{w_{ir}\}_{r\in\mc N_i}$,~$\mb{d}_0^i=\nabla f_{i,s_0^i}(\bth_0^i)$, where~$s_0^i$ is chosen uniformly at random in~$\{1,\cdots,m_i\}$
		\For{$k= 0,1,2,\cdots$}
	    \State\textbf{Update}~$\bds\theta_{k+1}^i = \sum_{r\in\mc{N}_i}w_{ir}\bds\theta_k^r- \alpha_k \mb{d}_k^i$
	    \State \textbf{Choose}~$s_{k+1}^i$ uniformly at random in~$\{1,\cdots,m_i\}$ 
		\State \textbf{Compute} the local stochastic gradient~$\nabla f_{i,s_{k+1}^i}(\bth_{k+1}^i)$
		\State \textbf{Update}:~$\mb{d}_{k+1}^i = \sum_{r\in\mc{N}_i}w_{ir}\mb{d}_{k}^r + \nabla f_{i,s_{k+1}^i}(\bth_{k+1}^i) - \nabla f_{i,s_{k}^i}(\bth_{k}^i)$
		\EndFor
	\end{algorithmic}
\end{algorithm}

\begin{rmk}[DSGD vs. GT-DSGD]\label{R4}
By comparing DSGD~\eqref{DSGD_convergence} and GT-DSGD~\eqref{DSGT_convergence}, we note that under a constant step-size, GT-DSGD removes the bias~$\mc{O}\left(\frac{\alpha^2\kappa^2b}{\left(1-\lambda\right)^2}\right)$ that comes from~$b \triangleq \frac{1}{n}\sum_{i=1}^{n}\left\|\nabla f_i\left(\bds\theta^*\right)\right\|^2$ in DSGD. However, the network dependence in GT-DSGD,~$\mc{O}\left(\frac{1}{(1-\lambda)^3}\right)$, is worse than DSGD where it is~$\mc{O}\left(\frac{1}{(1-\lambda)^2}\right)$. A tradeoff here is imminent where the two approaches have their own merits depending on the relative sizes of~$b$ and~$\lambda$. Clearly, when the bias~$b$ dominates, e.g., when the data across nodes is largely heterogeneous, GT-DSGD achieves a lower steady-state error than DSGD. 
Under diminishing
step-sizes, DSGD and GT-DSGD have comparable performance.
Of relevance here are EXTRA~\cite{EXTRA} and Exact Diffusion~\cite{Exact_Diffusion}, both of which eliminate the bias caused by~$b$ and are built on a different principle from gradient tracking.  
\end{rmk}

\begin{rmk}[SGD vs. GT-DSGD]\label{R5}
Note that with constant step-sizes, the performance of SGD in~\eqref{sgd_conv} and GT-DSGD in~\eqref{DSGT_convergence} is comparable. In particular, both methods converge linearly but there is a steady-state error, which is controlled by the step-size~$\alpha$ and the variance~$\sigma^2$ of the stochastic gradient, see Remark~\ref{R2}. Since GT-DSGD removes the bias in DSGD that comes due to the difference of the local and global objectives (see~$b$ in~\eqref{DSGD_convergence}), it may be considered as a more appropriate decentralization of SGD. This argument naturally leads to the idea that one can incorporate the centralized Variance Reduction (VR) techniques in the GT-DSGD to further improve the performance and achieve faster convergence. As we show in the following, adding variance reduction to GT-DSGD in fact leads to an exact linear convergence with a~\textit{constant} step-size and further improves its network dependence to~$\mc{O}\left(\frac{1}{(1-\lambda)^2}\right)$. 
\end{rmk}

\begin{rmk}[DSGD + VR]\label{dsgd_vr}
We emphasize that adding VR to DSGD does not enable exact linear convergence. Following Remark~\ref{VRdisc}, VR removes the steady-state error caused by the variance of the stochastic gradient. However, in a decentralized setting, the heterogeneity across the local data batches is not accounted for unless gradient tracking is employed. This difference between the local batches across the nodes is captured by the aforementioned bias~$b$ in~\eqref{DSGD_convergence} and is removed by gradient tracking that estimates the average of local gradients across the nodes.
\end{rmk}

\section{Decentralized Variance-Reduced Methods with Gradient Tracking}\label{S3}
We now provide a unified algorithmic framework, GT-VR, that provably improves DSGD and follows from Remarks~\ref{R5} and~\ref{dsgd_vr}. This framework combines variance-reduction with GT-DSGD to achieve both robust performance and fast convergence. First, recall from \textit{Section: Variance-Reduced Stochastic Gradient Descent} that VR methods iteratively estimate the batch
gradient from randomly drawn samples. In the decentralized case, each node~$i$ thus implements VR locally to estimate its local batch gradient~$\nabla f_i$. Gradient tracking, on the other hand, estimates the average of the local VR estimators across the nodes and can be thought of as fusion in space. 
Consequently, VR and gradient tracking jointly learn the global batch gradient~$\nabla F$ at each node asymptotically. For definiteness, we present and analyze two instances of GT-VR, namely, GT-SAGA and GT-SVRG, and show that they achieve exact linear convergence with constant step-sizes for the class of smooth and strongly-convex functions. We further show that in a ``big-data" regime, both GT-SAGA and GT-SVRG act effectively as means for parallel computation and achieve a linear speed-up compared with their centralized counterparts.

\vspace{-0.2cm}
\subsection{GT-SAGA}
To implement the SAGA estimators locally, each node~$i$ maintains a gradient table that stores all local component gradients~$\{\nabla f_{i,j}(\wh{\bth}_{i,j})\}_{j=1}^{m_i}$, where~$\wh{\bth}_{i,j}$ represents the most recent iterate where the gradient of~$f_{i,j}$ was evaluated. At iteration~$k\geq0$, each node~$i$ chooses an index~$s_k^i$ uniformly at random from~$\{1,\cdots,m_i\}$ and computes the local SAGA gradient~$\mb{g}_k^i$ as
\begin{align}
\mb{g}_{k}^{i} = \nabla f_{i,s_k^i}\big(\bth_{k}^{i}\big) - \nabla f_{i,s_k^i}\big(\wh{\bth}_{i,s_k^i}\big) + \frac{1}{m_i}\sum_{j=1}^{m_i}\nabla f_{i,j}\big(\wh{\bth}_{i,j}\big),
\end{align}
where it can be shown that~$\mb{g}_k^i$ is an unbiased estimator of the local batch gradient~$\nabla f_i(\bth_k^i)$. Next, the element~$\nabla f_{i,s_k^i}(\wh{\bth}_{i,s_k^i})$ in the gradient table is replaced by~$\nabla f_{i,s_k^i}\big(\bth_{k}^{i}\big)$, while the other elements remain unchanged. The gradient tracking iteration~$\mb d_k^i$ is then implemented on the estimators~$\mb g_k^i$'s. The complete implementation of GT-SAGA~\cite{GTVR} is summarized in Algorithm~\ref{GT-SAGA}. 
\begin{algorithm}
\caption{GT-SAGA at each node~$i$}
\label{GT-SAGA}
\begin{algorithmic}[1] \Require$\bds\theta_0^i$,~$\alpha$,~$\{w_{ir}\}_{r\in\mc N_i}$,~$\mb{d}_0^i=\mb{g}_0^i=\nabla f_i(\bth_0^i)$, Gradient table $\{\nabla f_{i,j}(\wh{\bth}_{i,j})\}_{j=1}^{m_i}$,~$\wh{\bth}_{i,j} = \bth_0^i,\forall j$.
\For{$k= 0,1,2,\cdots$}
\State\textbf{Update} {$\bth_{k+1}^i = \sum_{r\in\mc N_i}w_{ir}\bth_{k}^r - \alpha\mb{d}_{k}^{i}$;}
\State\textbf{Choose}~{$s_{k+1}^i$ uniformly at random from~$\{1,\cdots,m_i\}$;}
\State\textbf{Compute}~{$\mb{g}_{k+1}^i = \nabla f_{i,s_{k+1}^i}\big(\bth_{k+1}^{i}\big) - \nabla f_{i,s_{k+1}^i}\big(\wh{\bth}_{i,s_{k+1}^i}\big) + \frac{1}{m_i}\sum_{j=1}^{m_i}\nabla f_{i,j}\big(\wh{\bth}_{i,j}\big)$;} \label{saga}
\State{\textbf{Replace}~$\nabla f_{i,s_{k+1}^i}\big(\wh{\bth}_{i,s_{k+1}^i}\big)$ by $\nabla f_{i,s_{k+1}^i}\big(\bth_{k+1}^{i}\big)$ in the gradient table.}
\State{\textbf{Update}~$\mb{d}_{k+1}^{i} = \sum_{r\in\mc N_i}w_{ir}\mb{d}_{k}^{r} + \mb{g}_{k+1}^i - \mb{g}_k^{i}$;}
\EndFor
\end{algorithmic}
\end{algorithm}

Similar to centralized SAGA~\cite{SAGA}, GT-SAGA converges linearly to~$\bth^*$ with a constant step-size. More precisely, assuming each~$f_{i,j}\in\mc{S}_{\mu,L}$ and by choosing~$\alpha = \min\left\{\mc{O}\left(\frac{1}{\mu M}\right),\mc{O}\left(\frac{m}{M}\frac{(1-\lambda)^2}{L\kappa}\right)\right\}$, where~$m=\min_i\{m_i\},M=\max_i\{m_i\}$, we have~\cite{GTVR},
\begin{align}
\frac{1}{n}\sum_{i=1}^{n}\mathbb{E}\left[\left\|\bds\theta_k^i-\bds\theta^*\right\|_2^2\right] \leq R\left(1-\min\left\{\mc{O}\left(\frac{1}{M}\right),\mc{O}\left(\frac{m}{M}\frac{(1-\lambda)^2}{\kappa^2}\right)\right\}\right)^k,\qquad\forall k\geq 0,
\end{align}
for some~$R> 0$. In other words, GT-SAGA achieves~$\epsilon$-accuracy of~$\bth^*$ in 
$$\mc{O}\left(\max\left\{M,\frac{M}{m}\frac{\kappa^2}{(1-\lambda)^2}\right\}\log\frac{1}{\epsilon}\right)$$ parallel local component gradient computations. We emphasize that GT-SAGA, unlike the stochastic algorithms (DSGD and GT-DSGD) discussed before, exhibits linear convergence to the global minimizer~$\bth^*$ of~$F$. This exact linear convergence is a consequence of both variance reduction and gradient tracking; see Remarks~\ref{R6},~\ref{R7},~\ref{related_work} and~\ref{R8} for additional comments.

\vspace{-0.2cm}
\subsection{GT-SVRG}
GT-SVRG, formally described in Algorithm~\ref{GT-SVRG}, is a double-loop method, where the outer loop index is~$k$ and the inner loop index is~$t$, that builds upon the centralized SVRG. At every outer loop, each node~$i$ computes a local batch gradient and proceeds to a finite number~$T$ of inner loop iterations; in the inner loop, each node~$i$ performs GT-DSGD (type) iterations in addition to updating the local gradient estimate~$\mb v_t^i$ (Algorithm~\ref{GT-SVRG}: Step 7). It can be verified that~$\mb v_t^i$ is an unbiased estimator of the corresponding local batch gradient at node~$i$. In practice, all options (a)-(c) work similarly well. For example, under option (a), it is shown in~\cite{GTVR} that with~$\alpha=\mc{O}\left(\frac{(1-\lambda)^2}{L\kappa}\right)$ and~$T = \mc{O}\left(\frac{\kappa^2\log\kappa}{(1-\lambda)^2}\right)$, the outer loop of GT-SVRG follows:
\begin{align}
\frac{1}{n}\sum_{i=1}^{n}\mathbb{E}\left[\left\|\bds{\theta}_k^i-\bds\theta^*\right\|_2^2\right] \leq U\cdot0.9^k,   
\end{align}
for some~$U>0$. This argument implies that GT-SVRG achieves~$\epsilon$-accuracy of~$\bth^*$ in~$\mc{O}\left(\log\frac{1}{\epsilon}\right)$ outer loop iterations. We further note that each outer-loop update requires each node~$i$ to compute~$m_i+2T$ local component gradients. GT-SVRG thus achieves~$\epsilon$-accuracy of~$\bth^*$ in totally
$$
\mc{O}\left(\left(M + \frac{\kappa^2\log\kappa}{(1-\lambda)^2}\right)\log\frac{1}{\epsilon}\right)
$$
parallel local component gradient computations.

\begin{algorithm}
\caption{GT-SVRG at each node~$i$}
\label{GT-SVRG}
\begin{algorithmic}[1] 
\Require$\bds{\theta}_0^i$,~$\alpha$,~$\{w_{ir}\}_{r\in\mc N_i}$,~$\mb{d}_0^i = \mb{v}_0^i = \nabla f_i(\bth_0^i)$.
\For{$k= 0,1,2,\cdots$}
\State{\textbf{Initialize}~$\ul{\bth}_{0}^i = \bds{\theta}^{i}_{k}$}
\State{\textbf{Compute}~$\nabla f_i(\ul{\bth}_0^i)=\frac{1}{m_i}\sum_{j=1}^{m_i}\nabla f_{i,j}(\ul{\bth}_0^i)$ }
\For{$t= 0,1,2,\cdots,T-1$}
\State\textbf{Update} {$\ul{\bth}_{t+1}^i = \sum_{r\in\mc{N}_i}w_{ir}\ul{\bth}_{t}^r - \alpha\cdot\mb{d}_{t}^{i}$;}
\State\textbf{Choose}~{$s_{t+1}^i$ uniformly at random from~$\{1,\cdots,m_i\}$;}
\State\textbf{Compute}~{$\mb{v}_{t+1}^i = \nabla f_{i,s_{t+1}^i}\big(\ul{\bth}_{t+1}^i\big) - \nabla f_{i,s_{t+1}^i}\big(\ul{\bth}_0^i\big) + \nabla f_i(\ul{\bth}_0^i)$;} \label{saga}
\State{\textbf{Update}~$\mb{d}_{t+1}^i = \sum_{r\in\mc N_i}w_{ir}\mb{d}_{t}^{r} + \mb{v}_{t+1}^i - \mb{v}_{t}^i$;}
\EndFor
\State{\textbf{Set}~$\mb{d}_{0}^i = \mb{d}_{T}^i$} and~$\mb{v}_0^i = \mb{v}_T^i$
\State{Option (a):~\textbf{Set}~$\bds{\theta}_{k+1}^i = \ul\bth_{T}^i$ }
\State{Option (b):~\textbf{Set}~$\bds{\theta}_{k+1}^i = \frac{1}{T}\sum_{t=0}^{T-1}\ul\bth_{t}^i$ }
\State{Option (c):~\textbf{Set}~$\bds{\theta}_{k+1}^i$ as a random selection from~$\{\ul\bth_t^i\}_{t=0}^{T-1}$ }
\EndFor
\end{algorithmic}
\end{algorithm}

\begin{rmk}[GT-SAGA vs. GT-SVRG: Linear speedup]\label{R6}
Both GT-SAGA and GT-SVRG have a low per-iteration computation cost and converge  linearly to~$\bth^*$, i.e., they reach~$\epsilon$-accuracy of~$\bth^*$ respectively in~$\mc{O}\left(\max\left\{M,\frac{M}{m}\frac{\kappa^2}{(1-\lambda)^2}\right\}\log\frac{1}{\epsilon}\right)$ and~$\mc{O}\left(\left(M + \frac{\kappa^2\log \kappa}{(1-\lambda)^2}\right)\log\frac{1}{\epsilon}\right)$ parallel local component gradient computations. Interestingly, when the data sets at the nodes are large and balanced such that~$M\approx m \gg \frac{\kappa^2}{1-\lambda^2}$, the complexities of GT-SAGA and GT-SVRG become~$\mc{O}(M\log\frac{1}{\epsilon})$, independent of the network, and are~$n$ times faster than that of centralized SAGA and SVRG. Clearly, in this ``big-data" regime, GT-SAGA and GT-SVRG each acts effectively as a means for parallel computation and achieves a linear speed-up compared with its centralized counterpart. 
\end{rmk}

\begin{rmk}[GT-SAGA vs. GT-SVRG: Unbalanced data]\label{R7} It can also be observed that when data samples are distributed over the network in an unbalanced way, i.e.,~$\frac{M}{m}$ is large, GT-SVRG may achieve a lower complexity than GT-SAGA in terms of number of component gradient evaluations. However, from a practical implementation standpoint, an unbalanced data distribution may lead to a longer wall-clock time in GT-SVRG. This is because the next inner loop cannot be executed until all nodes finish their local batch gradient computations and nodes with a large amount of data take longer to finish this computation, leading to an overall increase in runtime. Clearly, there is an inherent trade-off between network synchrony, latency, and the storage of gradients as far as the relative implementation complexities of GT-SAGA and GT-SVRG are concerned. If each node is capable of storing all local component gradients, then GT-SAGA is preferable due to its flexibility of implementation and faster convergence in practice. On the other hand, for large-scale optimization problems where each node holds a very large number of data samples, storing all component gradients may be infeasible and therefore GT-SVRG may be preferred.
\end{rmk}

\begin{rmk}[Related work on decentralized VR methods]\label{related_work}
Existing decentralized VR methods include DSA~\cite{DSA} that combines EXTRA~\cite{EXTRA} with SAGA~\cite{SAGA}, diffusion-AVRG that combines exact diffusion~\cite{Exact_Diffusion} and AVRG~\cite{AVRG}, DSBA~\cite{DSBA} that adds proximal mapping~\cite{point-SAGA} to each iteration of DSA, ADFS~\cite{ADFS} that applies an accelerated randomized proximal coordinate gradient method~\cite{APCG} to the dual formulation of Problem P3, and Network-SVRG/SARAH~\cite{Network-DANE} that implements variance-reduction in the decentralized DANE framework based on gradient tracking. We note that in large-scale scenarios where~$M\approx m$ is very large, both~GT-SAGA and~GT-SVRG improve upon the convergence rate of these methods in terms of the joint dependence on~$\kappa$ and~$M\approx m$, with the exception of DSBA and ADFS. Both DSBA and ADFS achieve better iteration complexity, however, at the expense of computing the proximal mapping of a component function at each iteration.	Although the computation of this proximal mapping is efficient for certain function classes, it can be very expensive for general functions.
\end{rmk}

\begin{rmk}[Communication complexity]\label{R8}
We now compare the communication complexities of the decentralized algorithms discussed in this article. Since the node deployment is not necessarily deterministic, we provide the expected number of communication rounds per node required to achieve an~$\epsilon$-accurate solution (each communication is over a~$p$-dimensional vector). Note that DSGD, GT-DSGD, and GT-SAGA all incur~$\mc{O}(d_{\mbox{\footnotesize exp}})$ expected number of communication rounds per node, at each iteration, where~$d_{\mbox{\footnotesize exp}}$ is the expected degree of the (possibly random) communication graph~$\mc{G}$. Thus, their \textit{expected communication complexity} is their iteration complexity scaled by~$d_{\mbox{\footnotesize exp}}$ and is given by~$\mc{O}(d_{\mbox{\footnotesize exp}}\frac{1}{\epsilon})$,~$\mc{O}(d_{\mbox{\footnotesize exp}}\frac{1}{\epsilon})$, and~$\mc{O}\left(\max\left\{M,\frac{M}{m}\frac{\kappa^2}{(1-\lambda)^2}\right\}d_{\mbox{\footnotesize exp}}\log\frac{1}{\epsilon}\right)$, respectively. For GT-SVRG, we note that a total number of~$\mc{O}(\log\frac{1}{\epsilon})$ outer-loop iterations are required, where each corresponding inner loop incurs~$\mc{O}(T) = \mc{O}\left(\frac{\kappa^2\log\kappa}{(1-\lambda)^2}d_{\mbox{\footnotesize exp}}\right)$ rounds of communication, resulting into to a total communication complexity of~$\mc{O} \left( \frac{\kappa^2\log\kappa}{(1-\lambda)^2}d_{\mbox{\footnotesize exp}}\log\frac{1}{\epsilon} \right)$. Clearly, GT-SAGA and GT-SVRG, due to their fast linear convergence, improve upon the communication complexities of DSGD and GT-DSGD. It is further interesting to observe that in the big-data regime where each node has a large number of data samples, GT-SVRG achieves a lower communication complexity than GT-SAGA. Finally, we note that all gradient-tracking based algorithms require two consecutive rounds of communication per stochastic gradient evaluation with neighboring nodes to update the estimate~$\bth_k^i$ and the gradient tracker~$\mb{d}_k^i$, respectively. This may increase the communication burden of the network especially when~$\bth_k^i$ is of high dimension. We note that, for the sake of completeness, we add~$d_{\mbox{\footnotesize exp}}$ to the communication complexities, which is a function of the underlying graph~$\mc{G}$; in particular,~$d_{\mbox{\footnotesize exp}}\!=\!\mc{O}(1)$ for random geometric graphs (assuming constant density of deployment of nodes) and~$d_{\mbox{\footnotesize exp}}\!=\!\mc{O}(\log n)$ for exponential graphs, see also \textit{Section: Numerical Illustrations} on these graphs.
\end{rmk}

\section{Numerical Illustrations}\label{S5}
\begin{figure*}[!h]
\centering
\subfigure{\includegraphics[width=2in]{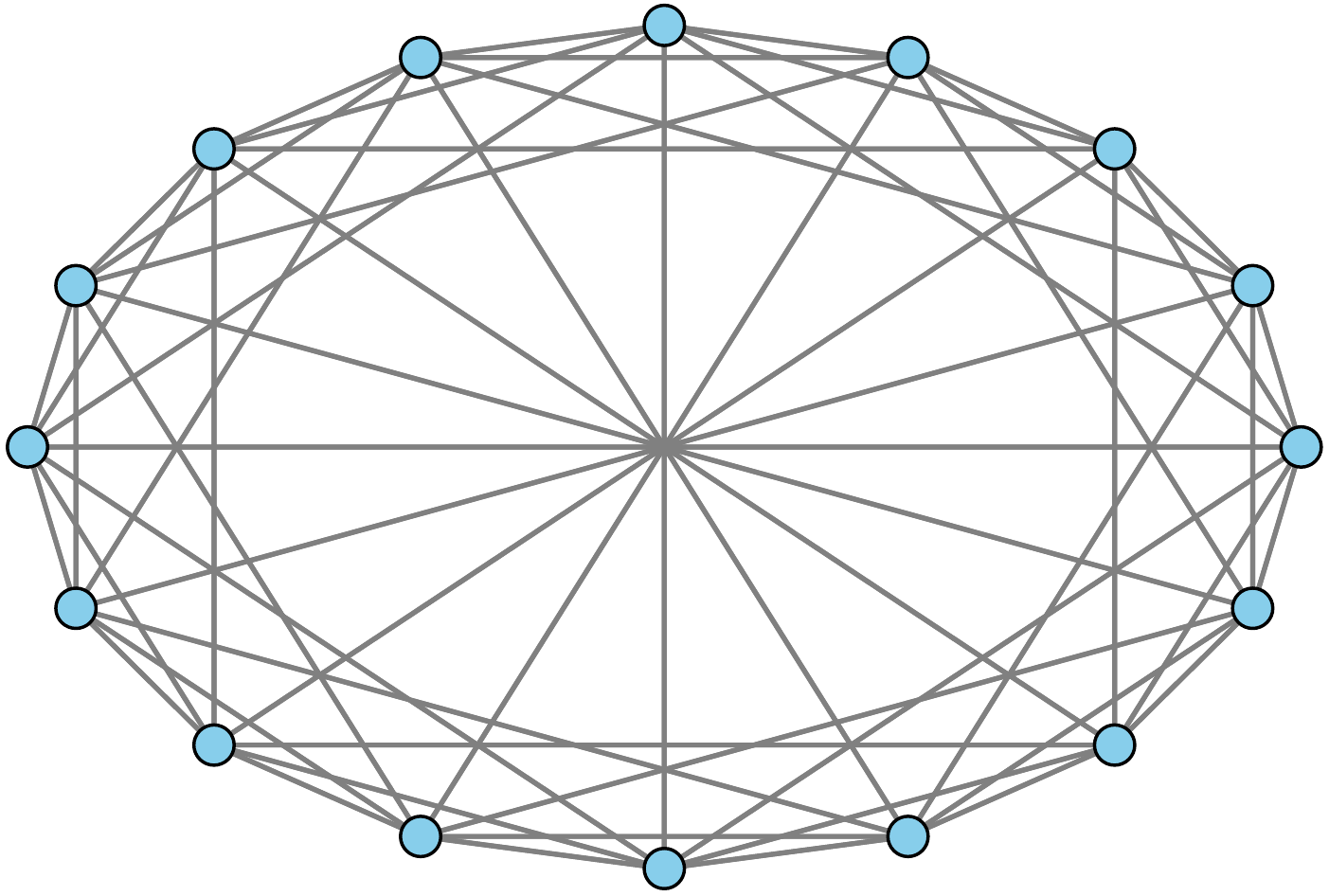}}\hspace{2cm}
\subfigure{\includegraphics[width=2in]{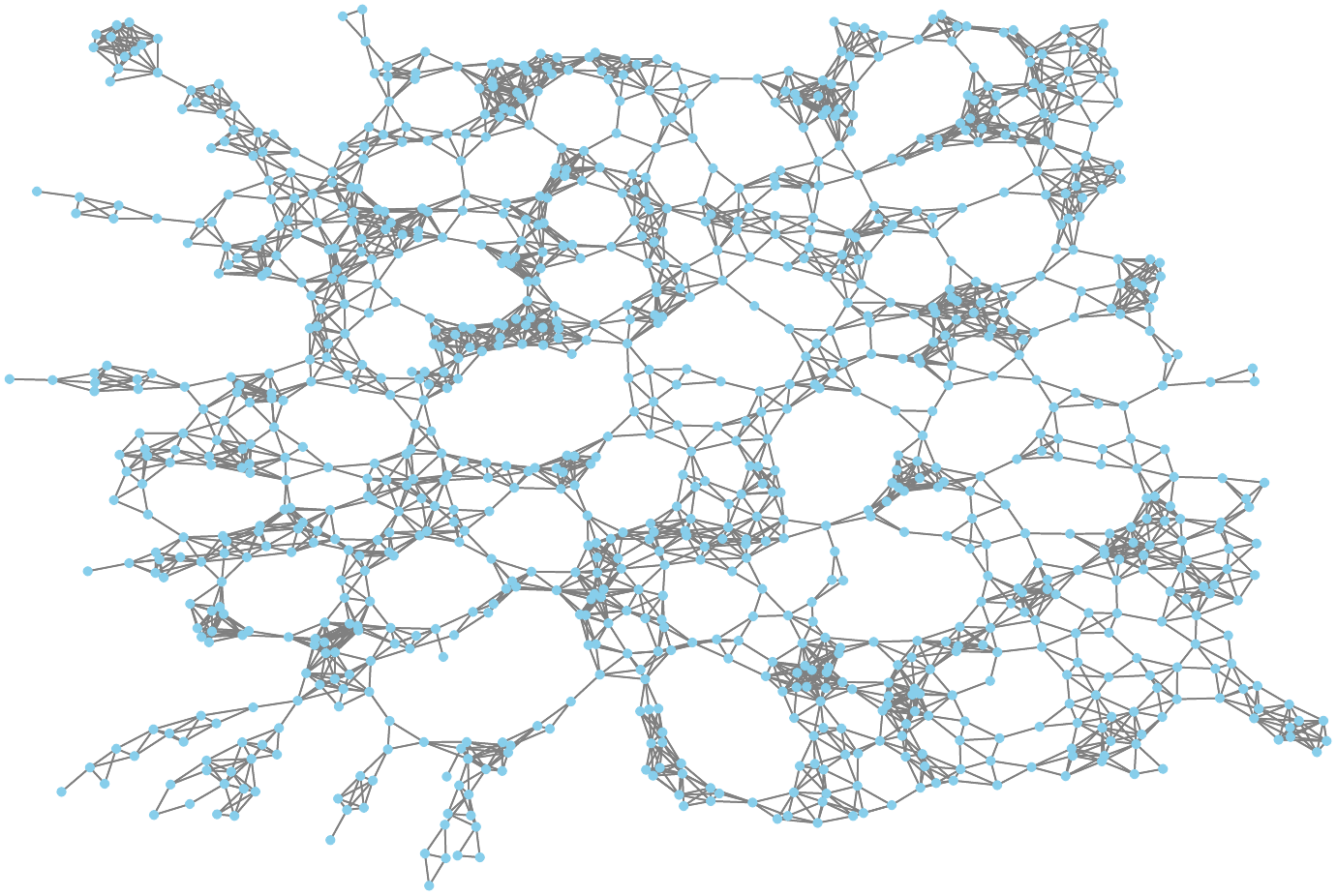}}
\caption{(Left) An exponential graph with~$16$ nodes. (Right) A random geometric graph with~$1,000$ nodes.}
\label{fig_graphs}\label{decentralized}
\end{figure*}

In this section, we present numerical experiments to illustrate the convergence properties of the decentralized stochastic optimization algorithms discussed in this article, i.e., DSGD, GT-DSGD, GT-SAGA, and GT-SVRG. We show experimental results on two different types of graphs shown in Fig.~\ref{fig_graphs}: 
\begin{inparaenum}[(i)]
\item an exponential graph with~$n=16$ nodes modeling a highly-structured training environment with a large number of data samples per node; and,
\item a random geometric graph with~$n=1,000$ nodes modeling a large-scale, ad-hoc training scenario. Their associated doubly-stochastic weight matrices~$W$ are generated by the Metroplis method with the second largest eigenvalue~$\lambda$ of~$0.75$ in the former and~$0.9994$ in the latter. 
\end{inparaenum} The decentralized training problem we consider is  classification of hand-written digits from the MNIST dataset~\cite{mnist} with the help of logistic regression (strongly-convex) and a two-layer neural network (non-convex). 

\vspace{-0.5cm}
\subsection{Logistic Regression: Strongly-convex}
We first compare the algorithms of interest in the context of training a regularized logistic regression model~\cite{PRML}, that is smooth and strongly-convex, to classify two digits~$\{3,8\}$. We use a total of~$N \!=\! 12,000$ images for training and~$1,966$ images for testing. Each node~$i$ holds~$m_i$ training samples, i.e.,~$\left\{\mb{x}_{i,j},y_{i,j}\right\}_{j=1}^{m_i}\subseteq\mathbb{R}^{784}\times\left\{-1,+1\right\}$, where~$\mb{x}_{i,j}$ is the feature vector (image) and~$y_{i,j}$ is the corresponding binary label. The nodes cooperate to solve the following problem:
	\begin{equation}
	\operatorname*{min}_{\mb{b}\in\mathbb{R}^{784},\:c\in\mathbb{R}}F(\mb{b},c) = \frac{1}{n}\sum_{i=1}^{n}\frac{1}{m_i}\sum_{j=1}^{m_i}\ln\left[1+\exp\left\{-(\mb{b}^\top\mb{x}_{i,j}+c)y_{i,j}\right\}\right]+\frac{\lambda}{2}\|\mb{b}\|_2^2, \nonumber
	\end{equation}
where~$\bth=[\mb b^\top c]^\top$, the regularization parameter is~$\lambda = 1/N$, and the features are normalized to unit vectors~\cite{SAG,DAVRG}. We plot the optimality gap, i.e.,~$F({\ol{\bth}}_k) - F(\bth^*)$, vs. the number of parallel component gradient evaluations and compare the algorithms in both balanced and unbalanced data distribution scenarios, recall Remarks~\ref{R6} and~\ref{R7}. The step-size for all algorithms is constant and is chosen to be~$1/L$, while the inner-loop length~$T$ of GT-SVRG is~$N/n$ in the case of balanced data and~$4N/n$ in the case of unbalanced data. 

\begin{figure*}[!h]
\centering
\subfigure{\includegraphics[width=5.90in]{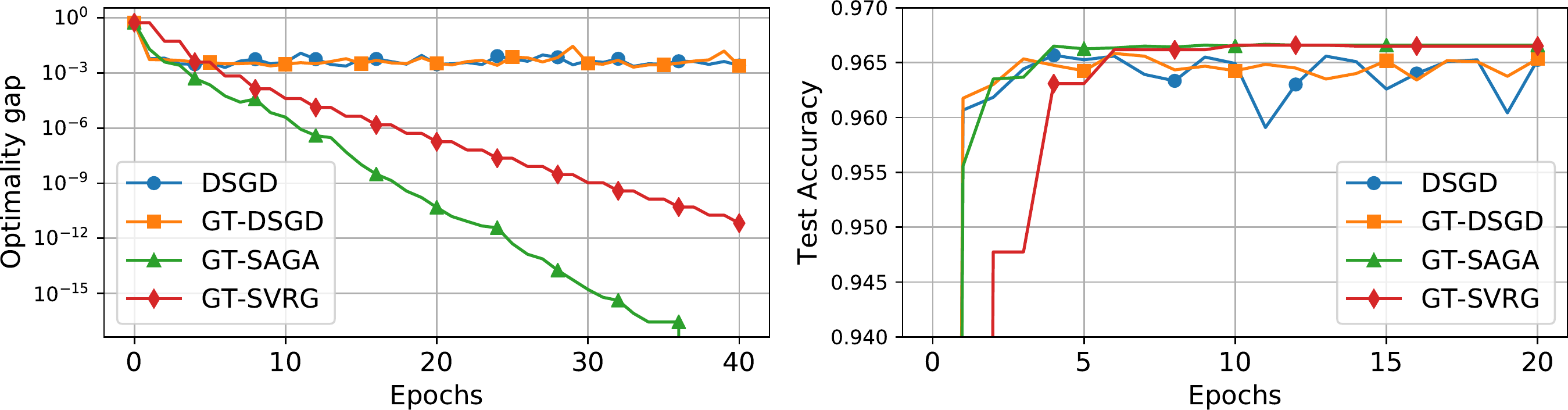}}
\caption{Decentralized logistic regression with balanced data over the~$16$-node exponential graph, where each epoch represents $N/n = 750$ component gradient evaluations at each node.}
\label{LR_balanced}
\end{figure*}

\textbf{Balanced Data:} To model a stable training environment with a balanced data distribution, e.g., in data centers or computing clusters, we choose a highly structured, well-connected, exponential graph with~$n=16$ nodes resulting into a relatively large number of samples~($m_i=750$) per~node. Each node has approximately the same number of images in each class, i.e., the data distribution is balanced and homogeneous, leading to similar local cost functions among the nodes and therefore the bias term~$b$ in DSGD is relatively small. From Remarks~\ref{R2} and~\ref{R4}, recall that when~$b$ is small and the graph is well-connected, DSGD and GT-DSGD exhibit similar performance that is also verified numerically in Fig.~\ref{LR_balanced}. Adding variance reduction to GT-DSGD however significantly improves the performance in terms of both the optimality gap and the test accuracy, leading to a linear convergence in both GT-SAGA and GT-SVRG to the exact solution.

\begin{figure*}[!h]
\centering
\subfigure{\includegraphics[width=5.90in]{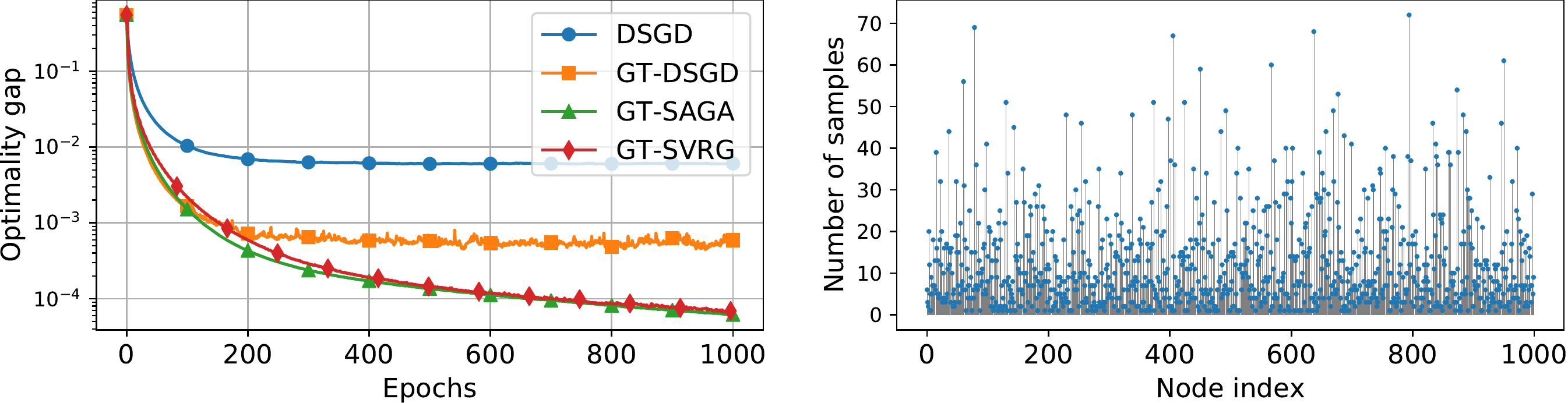}}
\caption{Decentralized logistic regression with unbalanced data over a~$1,000$-node random geometric graph, where each epoch represents~$N/n = 12$ component gradient evaluations at each node.}
\label{LR_unbalanced}
\end{figure*}

\textbf{Unbalanced Data:} We next compare the algorithms when the data distribution is unbalanced and the nodes interact over a random geometric graph of~$n=1,000$ nodes, modeling a large-scale, wireless communication network. In this case, the~$N = 12,000$ training images are randomly distributed among the nodes, see Fig.~\ref{LR_unbalanced} (right) for the number of training samples at each node. We make a further restriction that the training data samples at each node belong to only one class, either~$3$ or~$8$. This leads to unbalanced data sizes and heterogeneous data distributions at the nodes, making the local functions significantly different from each other and thus the bias~$b$ in DSGD is relatively large. The performance comparison is shown in Fig.~\ref{LR_unbalanced} (left), where it can be observed that DSGD degrades considerably in this case and the addition of gradient tracking results into a smaller steady-state error (Remark~\ref{R4}). Adding variance reduction, as before, leads to a linear convergence to the exact solution. 

\textbf{Discussion:} In both balanced and unbalanced data scenarios, the performance improvement due to gradient tracking comes at a price of one additional round of communication per iteration, see also Remark~\ref{R8}. The addition of variance reduction in GT-SAGA and GT-SVRG significantly outperforms both DSGD and GT-DSGD. Their linear convergence however comes at a price of additional storage in GT-SAGA and a synchronization overhead in GT-SVRG. From Remark~\ref{R6}, we recall that when each node has roughly the same number of training samples, GT-SAGA converges faster than GT-SVRG  in terms of the number of parallel component gradient computations required, as can be observed in Fig.~\ref{LR_balanced}. On the other hand, as discussed in Remark~\ref{R7}, the iteration complexity of GT-SVRG is more robust to unbalanced data as it is independent of the~$M/m$ factor that appears in GT-SAGA, as it is shown in Fig.~\ref{LR_unbalanced}, where GT-SAGA and GT-SVRG exhibit similar convergence. However, GT-SVRG may incur additional latency and synchronization when the data is unbalanced, due to the different computing time of the local batch gradient evaluations across the network, before the execution of each inner-loop. 

\vspace{-0.3cm}
\subsection{Neural Network: Non-convex}
\begin{figure*}[!h]
\centering
\subfigure{\includegraphics[width =5.9in]{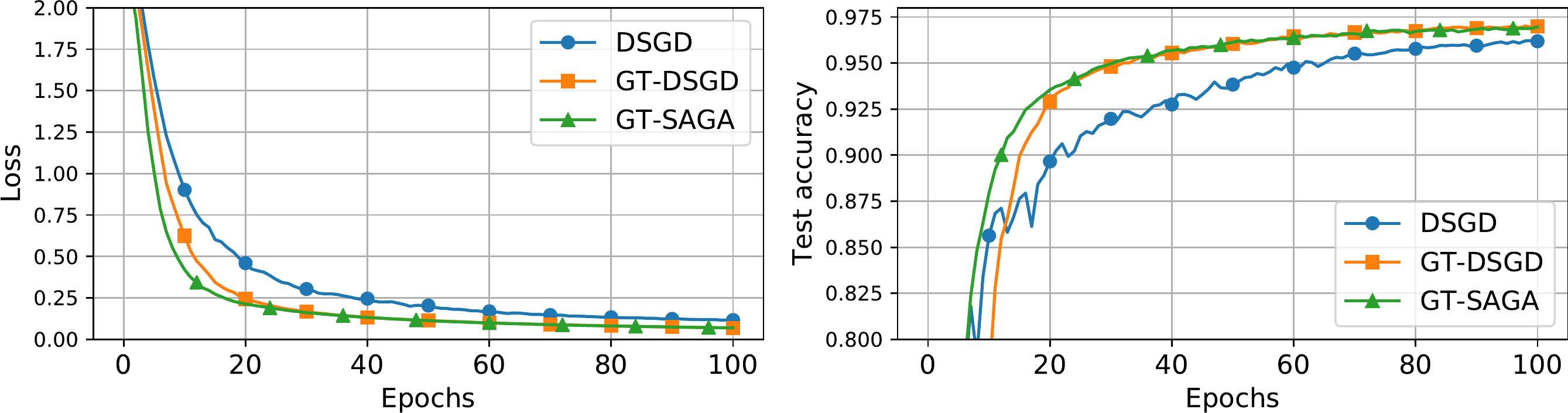}}
\caption{Two layer neural network over a~$1,000$-node random geometric graph, where one epoch represents~$N/n=60$ component gradient evaluations at each node.}
\label{LR_NN}
\end{figure*}

We now compare the performance of the algorithms when training a neural network with a non-convex loss function. The local neural network implemented at each node has one fully-connected hidden layer with~$64$ neurons and~$51,675$ parameters in total. The goal is to train a neural network that classifies all ten digits~$\{0,\ldots,9\}$ from the MNIST dataset with~$60,000$ training samples (around~$6,000$ images in each class) and~$10,000$ test images. The training dataset is divided randomly over~$1,000$ nodes such that each node has~$60$ data points. All algorithms use a constant step-size that is manually optimized for best performance.
Fig.~\ref{LR_NN} shows the loss~$F(\ol{\bth}_k)$ and the test accuracy over epochs. We note that adding gradient tracking to DSGD improves both the transient and steady-state performance in this non-convex setting. Similarly, adding variance-reduction improves the performance further. This behavior is also notable in the test accuracy. 
\vspace{-0.5cm}

\vspace{-0.5cm}
\section{Extensions and Discussion}\label{S4}
We now discuss some recent progress on several key aspects of decentralized optimization relevant to the first-order stochastic approaches described in this article. 

\textbf{Directed Graphs: }The methods described in this article are restricted to undirected graphs. Over~\textit{directed graphs}, the main challenge is that the weight matrices are either row-stochastic (RS) or column-stochastic (CS), but cannot be doubly-stochastic (DS), in general. A well-studied solution to this issue is based on the push-sum (type) algorithms~\cite{push_sum} that enable consensus with non-DS weights with the help of eigenvector estimation. Combining push-sum respectively with 
DSGD~\cite{DSGD_nedich,diffusion_Chen}, and~GT-DGD~\cite{GT_CDC,GT_Qu,DIGing} leads to SGP~\cite{SGP_ICML}, and ADD-OPT~\cite{ADD-OPT} that require CS weights. A similar idea is used in FROST~\cite{FROST} to implement decentralized optimization with RS weights. The issue with push-sum based extensions is that they require eigenvector estimation, which in itself is an iterative procedure and may slow down the underlying algorithms especially when the corresponding communication graphs are not well-connected. More recently, it is shown that GT-DGD~\eqref{DGT}, ADD-OPT, and FROST are special cases of the AB algorithm~\cite{AB,push-pull} that employs RS weights in~\eqref{DGT1} and CS weights in~\eqref{DGT2}, and thus is immediately applicable to arbitrary directed graphs. 
The AB framework naturally leads to stochastic optimization with gradient tracking over directed graphs, see SAB~\cite{SAB} that extends GT-DSGD to directed graphs, and further opens the possibility to extend GT-SAGA and GT-SVRG to their directed counterparts. 

\textbf{Communication and computation aspects: }Communication efficiency is an important aspect of decentralized optimization since communication can potentially become a bottleneck of the system when nodes are frequently transmitting high-dimensional vectors (model parameters) in the network. Different communication-efficient schemes~\cite{DOPT_lan, Network-DANE}, communication/computation tradeoffs~\cite{tutorial_nedich}, asynchronous implementations~\cite{Asyspa}, and quantization techniques~\cite{QDGD,QDGD2} have been studied with existing decentralized methods to efficiently manage the resources at each~node.

\textbf{Master-worker architectures:} The problems described in this article have experienced a significant research activity because of their direct applicability to large-scale training problems in machine learning~\cite{DGD_NIPS,SGP_ICML}. Since these applications are typically hosted in controlled settings, e.g., data centers with highly-sophisticated communication and a large number of highly-efficient computing clusters, master-worker architectures and parameter-server~models have become popular. In such architectures, see Fig.~\ref{decentralized} (left), a central master maintains the~current model parameters and communicates strategically with the workers, which individually hold a local batch of the total training data. 
Indeed, this architecture is not restricted to data centers alone and is also applicable to certain Internet-of-Things (IoT) scenarios where the devices are able to communicate to the master either directly via the cloud or via a mesh network among the devices. Various programming models and several variants of master-worker configurations have been proposed, such as MapReduce, All-Reduce, and federated learning~\cite{Federated_learning}, that are tailored for specific computing needs and environments. We emphasize that, on the contrary, the motivation behind the decentralized methods studied in this article comes from the scenarios where communication among the nodes is ad hoc, unstructured, and specialized topologies are not available. 

\section{Conclusions}\label{S6}
In this article, we discuss general formulation and solutions for decentralized, stochastic, first-order optimization methods. Our focus is on peer-to-peer networks that is applicable to ad-hoc wireless communication where the nodes have resource-constraints and limited communication capabilities. We discuss several fundamental algorithmic frameworks with a focus on gradient tracking and variance-reduction. For all algorithms, we provide a detailed discussion on their convergence rates, properties, and tradeoffs, with a particular emphasis on smooth and strongly-convex objective functions. An important line of future work in the field of decentralized machine learning is to analyze existing methods and develop new techniques for general non-convex objectives, given the tremendous success of deep neural networks.

\section*{Short Biographies}
\textbf{Ran Xin} (ranx@andrew.cmu.edu) is a PhD candidate in the Electrical and Computer Engineering (ECE) department at Carnegie Mellon University (CMU), PA. His research interests include optimization theory and methods. 

\textbf{Soummya Kar} (soummyak@andrew.cmu.edu) is an Associate Professor of ECE at Carnegie Mellon University, PA. His research interests include large-scale stochastic systems.

\textbf{Usman A. Khan} (khan@ece.tufts.edu) is an Associate Professor of ECE at Tufts University, MA. His research interests include optimization, control, and signal processing. 
		
\section*{Acknowledgments}
The authors acknowledge the support of NSF under awards CCF-1350264, CCF-1513936,  CMMI-1903972, and CBET-1935555. The authors would like to thank Boyue Li (CMU), Jianyu Wang (CMU), and Shuhua Yu (CMU) for their help and valuable discussions. 

\vspace{-0.2cm}
\bibliographystyle{IEEEbib}
\bibliography{SPM.bib}
\end{document}